\documentclass[11pt]{article}

\usepackage[preprint]{acl}
\usepackage{microtype}
\usepackage{times}
\usepackage{latexsym}
\usepackage{booktabs}
\usepackage{makecell}
\usepackage{amssymb}
\usepackage{multirow}
\usepackage{amsmath}
\usepackage{graphicx}
\usepackage{tikz}
\usetikzlibrary{positioning}
\usepackage[english,bidi=default]{babel} 

\title{StrategyBench: Evaluating Explicit Strategy Induction in Large Language Models}

\author{Jinghan Tan, Yuanzheng Wang, Lu Chen, \\
        Zijun Chen, Yuqian Wang, Maosong Sun}

\begin{document}

\maketitle
\begin{abstract}
As large language models are increasingly used in data-scarce and evolving task scenarios, few-shot in-context learning (ICL) has become a key paradigm for task adaptation. However, direct ICL often uses a small set of examples without explicitly abstracting task rules, making it sensitive to example construction. In contrast, human learners often reduce such sensitivity by first summarizing task rules from examples and then applying them to new instances. To evaluate this ability, we propose \textsc{StrategyBench}, which selects strategy-inducible tasks from BIG-Bench, constructs reference strategies, and defines evaluation metrics along two dimensions: strategy quality and downstream utility. We further analyze strategy induction from three perspectives: task variation, model configuration, and adaptation setting, covering category-wise differences, generator-executor choices, demonstration design, and SFT-based adaptation. Experiments show that explicit strategy utility differs substantially across task categories and depends on both strategy generation and execution conditions.
\end{abstract}

\section{Introduction}
Large language models have shown strong task adaptation ability under in-context learning (ICL)~\citep{brown2020language,wei2022chain,wang2023selfconsistency}. In standard few-shot ICL, which we refer to as direct ICL, models solve new tasks by conditioning on a small set of input-output examples. 
However, this paradigm often exploits examples only at a surface level and struggles to capture stable task rules. 
The limited context window restricts the number and diversity of examples, preventing models from observing sufficient example variations~\citep{agarwal2024many,liu2021makes,lu2022fantastically,min2022rethinking}. 
Moreover, even with more examples, models may rely on surface patterns or shortcuts rather than infer robust rules. 
Consequently, direct ICL remains sensitive to example construction and fragile under distribution shifts and complex reasoning scenarios~\citep{liu2024lost,mirzadeh2024gsm}.

\begin{figure}[!t]
\centering
\includegraphics[width=\columnwidth]{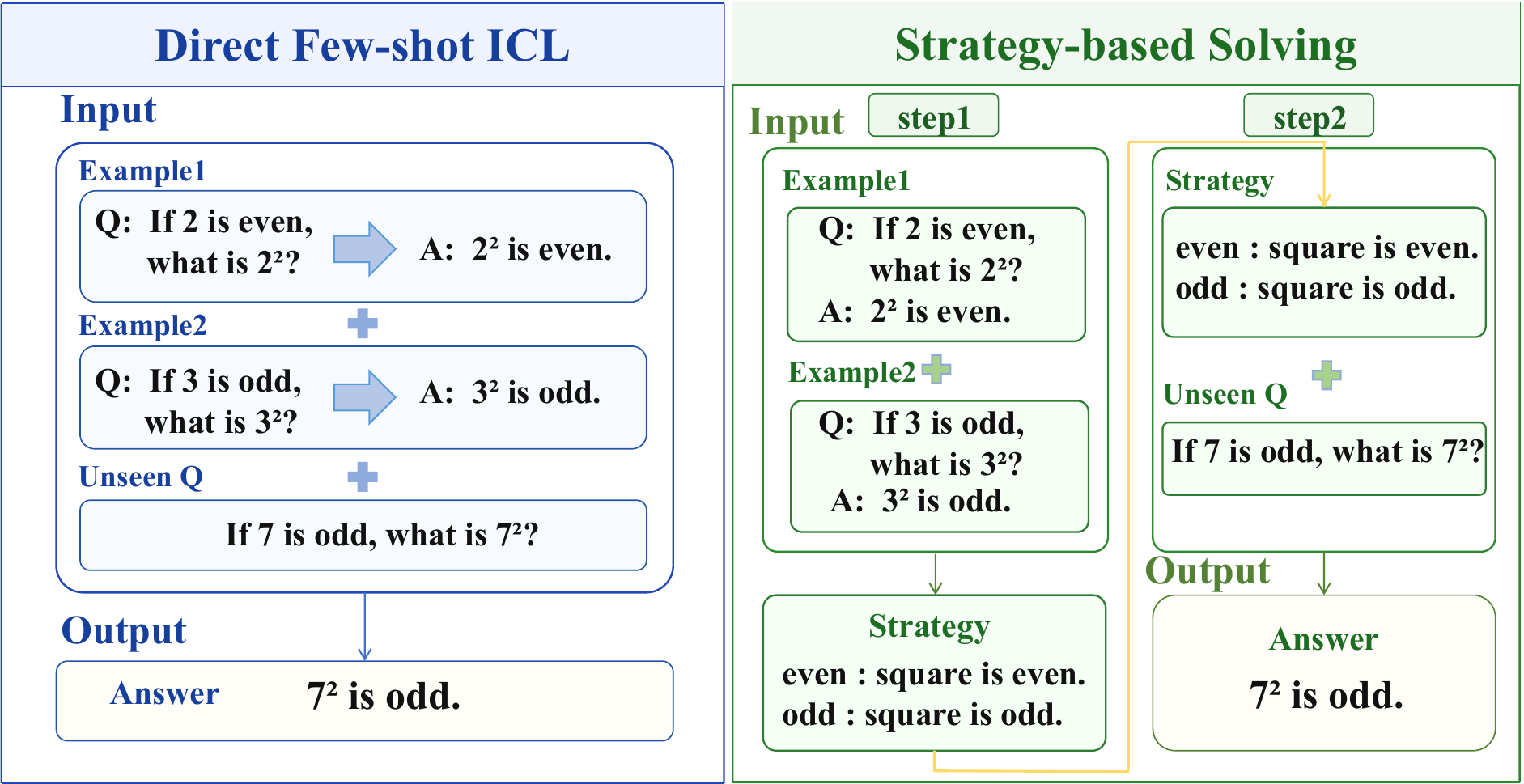}
\caption{Comparison between direct few-shot ICL and strategy-based solving. 
In direct few-shot ICL, the model directly predicts the answer for a new query
from the provided input-output examples. In strategy-based solving, the model
first induces an explicit task-level strategy from the few-shot examples and then
applies the induced strategy to solve the unseen query.}
\label{fig:direct_vs_strategy}
\vspace{-0.5em}
\end{figure}


To address these limitations, recent studies draw inspiration from the human learning process of ``summarizing before applying'' and abstract strategy-like representations, such as instructions, skills, or rules, from few-shot examples to guide subsequent problem solving~\citep{honovich2023instruction,zheng2024take,chen2024skill}. We refer to this paradigm as strategy-based ICL, where models first induce explicit task strategies and then apply them to unseen questions. Unlike Direct ICL, strategy-based ICL makes the underlying task rules explicit and reduces dependence on example construction. Figure~\ref{fig:direct_vs_strategy} illustrates how the two paradigms use example information differently.

Despite the potential of intermediate representations such as explicit strategies, systematic benchmarks for evaluating strategy induction in large language models remain limited. Existing evaluations mainly measure final-task performance in knowledge understanding, task solving, and mathematical reasoning~\citep{hendrycks2020measuring,srivastava2022beyond,suzgun2022challenging,cobbe2021training}, offering only partial insight into intermediate reasoning processes. Recent process-level evaluations have begun to address this gap, but they mostly assess instance-specific solution steps rather than general task-level strategies~\citep{lightman2023lets,zeng2023mrgsm8k,zheng2024take}. However, evaluating strategies is non-trivial, as it requires assessing not only whether a strategy is clear and abstract, but also whether it can be reliably applied by models and improve performance on unseen examples.These requirements place higher demands on benchmark construction and metric design.


To fill this gap, we propose \textsc{StrategyBench}, a benchmark for evaluating whether large language models can induce explicit task-level strategies from few-shot examples. \textsc{StrategyBench} selects strategy-inducible tasks from BIG-Bench and constructs high-quality reference strategies as training data. Based on this dataset, we design a multi-dimensional framework that assesses generated strategies from two perspectives: strategy text quality and downstream application utility. We further analyze how strategy induction and application are affected by model scale, few-shot example number, strategy format constraints, supervised fine-tuning, and strategy executors.

The main contributions of this paper are summarized as follows:
\begin{itemize}
    \item \textbf{Benchmark construction.} 
    We introduce \textsc{StrategyBench}, which is built from BIG-Bench to evaluate explicit strategy induction and supports both evaluation and training with reference strategies.
    
    \item \textbf{Evaluation framework.} 
    We design a multi-dimensional evaluation framework, which measures both strategy quality and downstream utility. 
    
    \item \textbf{Experimental analysis.} 
    We analyze how model scale, few-shot settings, strategy format constraints, supervised fine-tuning, and strategy executors affect strategy induction and application. Results show that readable strategies do not always yield better execution, and that strategy utility depends on the interaction among the strategy generator, executor, and format constraints.
\end{itemize}

\section{Preliminaries}

\subsection{Task and Episode Formulation}

This section formalizes the task and episode setting and introduces the direct ICL and strategy-based ICL paradigms adopted in this work. Let $\mathcal{T}$ denote the evaluation task set. For each task $T \in \mathcal{T}$, an example is represented as $(q,a)\sim T$, where $q$ is the question and $a$ is the gold answer. Each task defines a class of problems that share the same input--output format and are governed by common underlying rules. For example, in a Chinese-remainder-theorem task, all examples require solving modular constraints.

In each few-shot episode, the model observes a demonstration set
\begin{equation}
\mathcal{D}^{T}_{\mathrm{few}}
=
\{(q_i,a_i)\}_{i=1}^{k},
\end{equation}
and is evaluated on an unseen example set
\begin{equation}
\mathcal{D}^{T}_{\mathrm{new}}
=
\{(q_j^{\mathrm{new}},a_j^{\mathrm{new}})\}_{j=1}^{N}.
\end{equation}

\paragraph{Direct ICL.}
Direct ICL predicts each unseen answer directly from the demonstrations and the query:
\begin{equation}
a_j =
h_{\theta}(\mathcal{D}^{T}_{\mathrm{few}}, q_j^{\mathrm{new}}),
\end{equation}
where $h_{\theta}$ is the task-solving model and $a_j$ is the prediction for $q_j^{\mathrm{new}}$.

\paragraph{Strategy-based ICL.}
Strategy-based ICL separates task solving into strategy induction and strategy application. A generator first induces a task-level strategy from the demonstrations:
\begin{equation}
s_T = f_{\theta_g}(\mathcal{D}^{T}_{\mathrm{few}}),
\end{equation}
where $s_T$ is the explicit strategy and $f_{\theta_g}$ is the strategy generator. An executor then predicts the answer using the strategy and the unseen query:
\begin{equation}
a_j =
g_{\theta_e}(q_j^{\mathrm{new}}, s_T),
\end{equation}
where $g_{\theta_e}$ is the strategy executor.





\section{Benchmark Construction}

\begin{figure*}[t]
\centering
\includegraphics[width=0.95\textwidth]{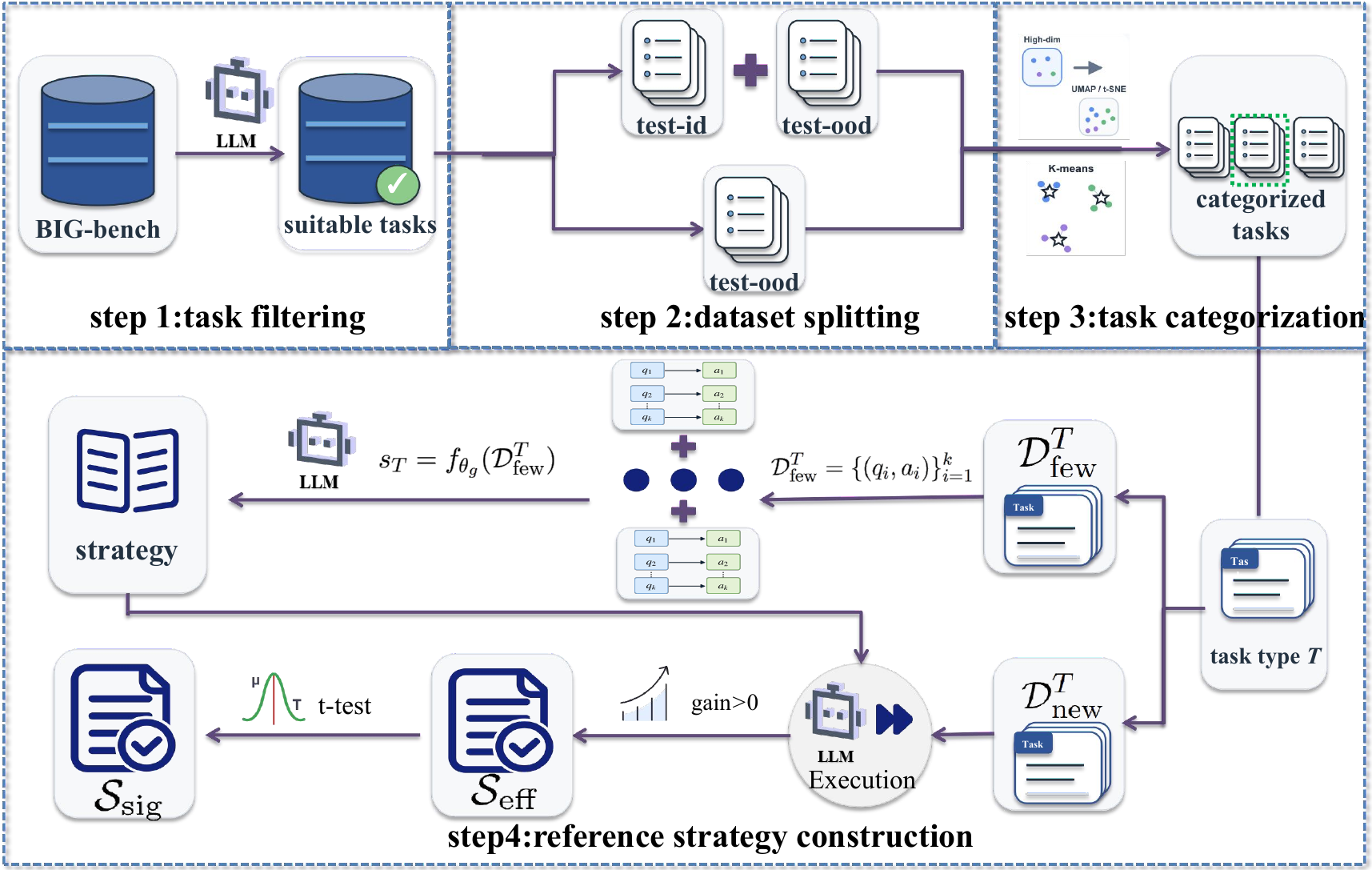}
\caption{Overall data construction pipeline of \textsc{StrategyBench}. Starting from BIG-Bench, we first filter tasks with the assistance of LLM-based pre-screening and human verification to obtain a suitable task pool. The filtered tasks are then processed along two independent dimensions: dataset splitting, which produces training, ID test, and OOD test subsets, and task categorization, which assigns tasks to coarse-grained categories. Based on the resulting task types and few-shot examples, candidate strategies are generated and further filtered by downstream utility and statistical significance to form the effective strategy set $\mathcal{S}_{\mathrm{eff}}$ and the significant reference strategy set $\mathcal{S}_{\mathrm{sig}}$.}
\label{fig:strategybench_construction_pipeline}
\end{figure*}

We build \textsc{StrategyBench} from BIG-Bench, which covers diverse task types such as commonsense reasoning, mathematical computation, language understanding, instruction following, and symbolic manipulation. Because BIG-Bench tasks differ substantially in generation methods, evaluation protocols, and input-output formats, we first perform task filtering, dataset splitting, and task categorization, and then construct the reference strategy set. Figure~\ref{fig:strategybench_construction_pipeline} illustrates the overall data construction pipeline.

\subsection{Task Filtering \& Data Normalization}



We follow the BBH-style procedure for task filtering and normalization. 
For filtering, we first remove tasks with dynamic evaluation, strong subjectivity, or overly long inputs. 
We then select strategy-inducible tasks through a ``model-assisted pre-filtering + human verification'' procedure along five dimensions: objective scorability, format stability, rule inductibility, low dependence on external knowledge, and strategy application potential. 
As shown in Table~\ref{tab:task_selection_statistics}, the filtering process substantially reduces the original candidate pool while retaining a diverse set of strategy-inducible tasks.

For normalization, we merge scattered task information into a unified input field and convert original answers into a unified target field. 
Detailed filtering rules and normalization procedures are provided in Appendix~\ref{app:task_selection}.

\begin{table}[t]
\centering
\footnotesize
\setlength{\tabcolsep}{3.5pt}
\renewcommand{\arraystretch}{1.00}
\begin{tabular*}{\columnwidth}{@{\extracolsep{\fill}}lccc}
\toprule
\textbf{Stage} & \textbf{\#Tasks} & \textbf{\#Subtasks} & \textbf{\#Examples} \\
\midrule
Before filtering & 108 & 581 & 1,614,716 \\
After filtering  & 44  & 356 & 123,845 \\
\bottomrule
\end{tabular*}
\caption{Dataset statistics before and after task filtering.}
\label{tab:task_selection_statistics}
\end{table}

\subsection{Dataset Splitting}

After filtering, we split the data into training and test sets. The test set is further divided into in-distribution and out-of-distribution subsets to evaluate strategy induction and generalization under different distribution shifts. We also retain BBH as an additional challenging test set. Detailed split definitions and rules are provided in Appendix~\ref{app:dataset_splitting}.
\subsection{Task Categorization}


After task filtering, we categorize the retained tasks to characterize their required abilities. We use an ``unsupervised clustering + human verification'' procedure and group tasks into six categories: \textbf{Numerical}, \textbf{Logic}, \textbf{Language}, \textbf{Spatial}, \textbf{Procedural}, and \textbf{Induction}. Unlike the original BIG-Bench labels, our categorization reorganizes tasks by required abilities, input-output patterns, and underlying solving rules. Details on semantic representation construction, clustering, and human verification are provided in Appendix~\ref{app:task_categorization}.

\subsection{Reference Strategy Construction}

We construct reference strategy sets to evaluate explicit strategies and support subsequent SFT experiments. The construction process includes two stages: candidate strategy generation and utility-driven filtering.

We first use gemini-3-flash-preview to generate candidate strategies for tasks in each category. We then compare their execution performance on unseen examples against direct ICL and retain strategies that improve performance as the effective strategy set $\mathcal{S}_{\mathrm{eff}}$. From this set, we further select strategies with statistically significant gains to form the significant reference strategy set $\mathcal{S}_{\mathrm{sig}}$. Table~\ref{tab:strategy_set_statistics} reports the statistics of both sets. Detailed generation settings, filtering rules, and additional statistics are provided in Appendix~\ref{app:strategy_construction}.

\begin{table}[t]
\centering
\small
\setlength{\tabcolsep}{6pt}
\renewcommand{\arraystretch}{1.08}
\resizebox{\columnwidth}{!}{
\begin{tabular}{lccc}
\toprule
\textbf{Strategy Set} & \textbf{\#Tasks} & \textbf{\#Subtasks} & \textbf{\#Strategies} \\
\midrule
$\mathcal{S}_{\mathrm{eff}}$ & 22 & 241 & 11,993 \\
$\mathcal{S}_{\mathrm{sig}}$ & 21 & 221 & 7,801 \\
\bottomrule
\end{tabular}
}
\caption{Statistics of the constructed reference strategy sets. 
$\mathcal{S}_{\mathrm{eff}}$ contains strategies that improve over Direct ICL, while $\mathcal{S}_{\mathrm{sig}}$ further retains strategies with statistically significant gains.}
\label{tab:strategy_set_statistics}
\end{table}

\section{Evaluation Metrics}


We build a multi-dimensional evaluation framework covering two perspectives: strategy application utility and strategy content quality. The former measures whether a strategy can be effectively applied to improve performance, while the latter evaluates the readability of the generated strategy.

\subsection{Strategy Utility}

For strategy application utility, we evaluate generated strategies from three aspects: correctness, stability, and answer efficiency. These metrics characterize the practical utility of strategies on unseen examples.
\paragraph{Correctness.}

Correctness measures whether a strategy helps the model answer unseen examples correctly. We use Corr as the main metric. Let $a_j^{(m)}(s_T)$ denote the prediction in the $m$-th sampling run under strategy $s_T$, with $M=3$ by default. We define:
{\small
\begin{equation}
\mathrm{Corr}_{j}(s_T)
=
\frac{1}{3}
\sum_{m=1}^{3}
\mathbf{1}\{a_{j}^{(m)}(s_T)=a_j^{\mathrm{new}}\}.
\end{equation}
}
Corr computes the fraction of predictions that match the gold answer across three sampling runs for the same query. A higher value indicates that the strategy more reliably guides the model to produce correct answers under repeated sampling.



\paragraph{Stability.}
Stability measures whether the model produces the same results when answering the same question multiple times (regardless of correctness). We evaluate it from two perspectives: sampling consistency $\mathrm{Sta}_{\mathrm{samp}}$, which measures consistency across repeated sampling runs, and robustness consistency $\mathrm{Sta}_{\mathrm{rob}}$, which measures consistency under perturbations of few-shot example order.

For sampling consistency, let $\{a_{j}^{(m)}(s_T)\}_{m=1}^{3}$ denote the three sampled answers for the $j$-th unseen example under strategy $s_T$, and let $v$ denote a candidate answer. We define:
{\small
\begin{equation}
\mathrm{Sta}_{\mathrm{samp}}^{j}(s_T)
=
\max_{v}
\frac{1}{3}
\sum_{m=1}^{3}
\mathbf{1}\{a_{j}^{(m)}(s_T)=v\}.
\end{equation}
}
This metric computes the fraction of the most frequent prediction across three sampling runs for the same query. A higher value indicates that the strategy more consistently guides the model to produce the same answer under repeated sampling.

For robustness consistency, let $s_T^{(m)}=f_{\theta_g}(\pi_m(\mathcal{D}^{T}_{\mathrm{few}}))$ denote the strategy generated after the $m$-th permutation of the few-shot examples, and let $a_j^{(m)}$ denote the corresponding prediction. We define:
\begin{equation}
\mathrm{Sta}_{\mathrm{rob}}^{j}
=
\max_{v}
\frac{1}{3}
\sum_{m=1}^{3}
\mathbf{1}
\left(
a_{j}^{(m)}=v
\right).
\end{equation}
This metric computes the fraction of the most frequent prediction across three random example-order perturbations for the same query. A higher value indicates that the strategy more robustly guides the model to produce consistent answers under such perturbations.

\paragraph{Efficiency.}


Answer efficiency measures whether a method can obtain correct answers with a lower token cost. Let $c_j(s_T)\in[0,1]$ denote the correctness score of strategy $s_T$ on the $j$-th query example. Let $l^{\mathrm{in}}_j(s_T)$ and $l^{\mathrm{out}}_j(s_T)$ denote the numbers of input and output tokens, respectively. Given the input length budget $L_{\mathrm{in}}^{\mathrm{budget}}$ and the output length budget $L_{\mathrm{out}}^{\mathrm{budget}}$, we define input and output length efficiency as:
{\small
\begin{equation}
\mathrm{LenEff}^{j}_{\mathrm{out}}(s_T)
=
c_j(s_T)
\cdot
\min\!\left(
\frac{L_{\mathrm{out}}^{\mathrm{budget}}}{l^{\mathrm{out}}_j(s_T)},
1
\right),
\end{equation}
}
{\small
\begin{equation}
\mathrm{LenEff}^{j}_{\mathrm{in}}(s_T)
=
c_j(s_T)
\cdot
\min\!\left(
\frac{L_{\mathrm{in}}^{\mathrm{budget}}}{l^{\mathrm{in}}_j(s_T)},
1
\right).
\end{equation}
}

If the prediction is incorrect, the efficiency score is 0. If the prediction is correct and the token cost does not exceed the budget, the method receives a high score. If the prediction is correct but requires a large token cost, the score is discounted according to the length ratio.

\subsection{Strategy Content Quality}

Strategy content quality assesses the generated strategy text itself, independent of its effectiveness in specific tasks. Here, we evaluate two metrics: Conciseness and Format Compliance.

\paragraph{ Conciseness.}
Conciseness measures repetition and redundancy within a strategy. Let $s_T=\{u_1,\dots,u_m\}$ denote a strategy, where $u_i$ is the $i$-th sentence, and let $\mathrm{sim}(u_i,u_j)$ denote the semantic similarity between two sentences. For $m\geq 2$,  we define Conc. as:
{\small
\begin{equation}
\mathrm{Conc}(s_T)
=
1-
\frac{2}{m(m-1)}
\sum_{i<j}\mathrm{sim}(u_i,u_j).
\end{equation}
}
When $m<2$, we set $\mathrm{Conc}(s_T)=1$. This formula computes the conciseness score by penalizing repeated semantic content within a strategy. A higher value indicates less internal repetition and more concise strategy text.

\paragraph{Format Compliance.}
Format compliance measures whether the model output follows the predefined
strategy format specified in the prompt.
An output is assigned a score of $1$ if it contains a complete strategy field
that satisfies the required parsing format, and $0$ otherwise:
{\small
\begin{equation}
\mathrm{Fmat}(s_T)
=
\mathbf{1}\{\text{valid strategy format in } s_T\}.
\end{equation}
}
where $\mathbf{1}[\cdot]$ is the indicator function. This metric is used to analyze whether the model can
consistently generate strategy outputs that satisfy the format constraints
required for automatic parsing. A higher Fmat. indicates that the model produces
more format-compliant strategy outputs.

\section{Experimental Setup}

\subsection{Baseline}


We compare three few-shot ICL baselines. \textbf{Direct ICL} solves unseen questions directly based on few-shot examples~\citep{brown2020language}. \textbf{CoT-prompted ICL} adds step-by-step reasoning instructions during inference~\citep{kojima2022large}. \textbf{Least-to-Most Prompting} decomposes a problem into subproblems and solves them sequentially~\citep{zhou2023least}. These methods represent three paradigms: example-based prediction, implicit reasoning-enhanced answer generation, and explicit intermediate reasoning.

\subsection{Experimental Configuration}


We use Qwen3-1.7B, 4B, 8B, and 14B as backbone models. Unless otherwise specified, Qwen3-8B serves as the default strategy generator and Qwen3-4B as the default executor. Prompting-based inference uses $k=3$ few-shot examples. For stability evaluation, we sample each query $3$ times and apply $M=3$ random perturbations to the few-shot example order. For fine-tuning, we construct SFT data from few-shot examples and reference strategies, and apply LoRA for parameter-efficient training. Detailed decoding parameters, generation and application settings, and training hyperparameters are provided in Appendix~\ref{app:implementation_details}.

\begin{table*}[!t]
\centering
\scriptsize
\setlength{\tabcolsep}{4.5pt}
\renewcommand{\arraystretch}{0.92}
\setlength{\aboverulesep}{0.35ex}
\setlength{\belowrulesep}{0.35ex}
\resizebox{\textwidth}{!}{
\begin{tabular}{llcccccc|c}
\toprule
\multirow{2}{*}{\textbf{Metric}} 
& \multirow{2}{*}{\textbf{Method}}
& \multicolumn{6}{c|}{\textbf{Task Category}}
& \multirow{2}{*}{\textbf{BBH}} \\
\cmidrule(l){3-8}
&
& \textbf{Numerical}
& \textbf{Logic}
& \textbf{Language}
& \textbf{Spatial}
& \textbf{Procedural}
& \multicolumn{1}{c|}{\textbf{Induction}}
&  \\
\midrule

\multirow{5}{*}{Corr.}
& L2M        & 58.43 & 42.34 & 22.21 & 6.21 & 9.47 & 27.50 & 24.09 \\
\cmidrule[0.2pt](lr){2-9}
& Direct ICL & 34.31 & 11.18 & 4.40 & 0.69 & 0.76 & 0.55 & 11.55 \\
& Ours       & 15.90 & 38.58 & 22.11 & 3.59 & 0.97 & 17.03 & 15.34 \\
\cmidrule[0.2pt](lr){2-9}
& CoT        & 57.30 & 43.12 & 19.43 & 4.71 & 19.91 & 28.99 & 21.40 \\
& CoT+Ours   & \textbf{69.58} & \textbf{55.32} & \textbf{36.09} & \textbf{8.62} & \textbf{25.95} & \textbf{40.90} & \textbf{37.79} \\
\midrule

\multirow{5}{*}{Sta$_{\mathrm{samp}}$}
& L2M        & 72.25 & 64.95 & 49.05 & 36.90 & 49.23 & 49.81 & 48.06 \\
\cmidrule[0.2pt](lr){2-9}
& Direct ICL & 73.27 & 39.80 & 39.47 & 34.67 & 34.95 & 35.60 & 44.73 \\
& Ours       & 60.15 & 68.53 & 52.10 & 36.24 & 54.84 & 48.90 & 51.90 \\
\cmidrule[0.2pt](lr){2-9}
& CoT        & 73.00 & 60.98 & 47.63 & 36.34 & 49.76 & 50.59 & 45.93 \\
& CoT+Ours   & \textbf{82.96} & \textbf{75.18} & \textbf{59.41} & \textbf{39.07} & \textbf{56.37} & \textbf{61.29} & \textbf{60.01} \\
\midrule

\multirow{5}{*}{Sta$_{\mathrm{rob}}$}
& L2M        & 77.82 & 72.41 & 60.83 & 52.84 & 60.61 & 60.03 & 59.89 \\
\cmidrule[0.2pt](lr){2-9}
& Direct ICL & 78.13 & 54.27 & 54.02 & 50.87 & 51.79 & 51.18 & 57.66 \\
& Ours       & 71.39 & 75.54 & 62.28 & 51.55 & \textbf{66.91} & 62.44 & 62.31 \\
\cmidrule[0.2pt](lr){2-9}
& CoT        & 78.67 & 69.69 & 59.65 & 52.23 & 61.48 & 61.14 & 58.54 \\
& CoT+Ours   & \textbf{86.40} & \textbf{80.16} & \textbf{67.89} & \textbf{53.96} & 66.70 & \textbf{70.48} & \textbf{68.91} \\
\midrule

\multirow{5}{*}{LenEff$_{\mathrm{in}}$}
& L2M        & 58.33 & 41.78 & 20.77 & 6.21 & 9.37 & 27.45 & 23.46 \\
\cmidrule[0.2pt](lr){2-9}
& Direct ICL & 34.21 & 11.05 & 4.20 & 0.66 & 0.75 & 0.52 & 11.53 \\
& Ours       & 15.89 & 38.22 & 20.89 & 3.58 & 0.97 & 17.03 & 15.21 \\
\cmidrule[0.2pt](lr){2-9}
& CoT        & 57.28 & 42.77 & 18.53 & 4.71 & 19.91 & 28.97 & 21.03 \\
& CoT+Ours   & \textbf{69.49} & \textbf{54.78} & \textbf{32.80} & \textbf{8.62} & \textbf{25.95} & \textbf{40.87} & \textbf{37.13} \\
\midrule

\multirow{5}{*}{LenEff$_{\mathrm{out}}$}
& L2M        & 23.47 & 18.50 & 9.09 & 2.51 & 3.85 & 10.70 & 9.58 \\
\cmidrule[0.2pt](lr){2-9}
& Direct ICL & 33.77 & 11.16 & 4.38 & 0.63 & 0.71 & 0.54 & 11.43 \\
& Ours       & 15.87 & 38.57 & 22.10 & 3.58 & 0.96 & 17.02 & 15.32 \\
\cmidrule[0.2pt](lr){2-9}
& CoT        & 37.55 & 33.86 & 13.40 & 3.30 & 13.46 & 17.93 & 14.01 \\
& CoT+Ours   & \textbf{56.04} & \textbf{49.00} & \textbf{30.19} & \textbf{6.76} & \textbf{22.69} & \textbf{35.66} & \textbf{29.49} \\

\bottomrule
\end{tabular}
}
\caption{Utility comparison across different task categories and BBH. Results are computed over all tasks within each category. \textbf{Bold} values indicate the best performance among the five methods for each metric within the same task category or BBH setting.}
\label{tab:utility_category_full_transposed}
\end{table*}

\section{Results and Analysis}

This section analyzes explicit strategy induction on \textsc{StrategyBench} through three research questions:

\begin{itemize}
    \item \textbf{RQ1: How do explicit strategies affect performance across task categories?}
    \item \textbf{RQ2: What factors affect strategy generation and application?}
    \item \textbf{RQ3: How does SFT affect strategy generation capability?}
\end{itemize}

\subsection {Main Result}

To answer \textbf{RQ1}, we compare our method with different baselines across task categories. We merge the ID and OOD test examples and compute results by task category. We also evaluate all methods on BBH. Table~\ref{tab:utility_category_full_transposed} reports the results.

Overall, CoT+Ours achieves the best results on most task categories and metrics, with clear gains on Numerical, Logic, and Language tasks. This suggests that explicit strategies and step-by-step reasoning prompts are complementary. Explicit strategies provide task-level rule constraints, while CoT strengthens instance-level reasoning. The strong results of CoT+Ours on Sta$_{\mathrm{samp}}$, Sta$_{\mathrm{rob}}$, LenEff$_{\mathrm{in}}$, and LenEff$_{\mathrm{out}}$ further show that this combination improves both output stability and answer efficiency.

Ours alone does not always outperform CoT or L2M, especially on Numerical and Procedural tasks. This suggests that explicit strategies serve better as high-level constraints and cannot fully replace instance-level reasoning or multi-step execution. When combined with CoT, the model can use both task-level rules and step-by-step reasoning, leading to more stable gains across most task categories.

\begin{table*}[t]
\centering
\footnotesize
\setlength{\tabcolsep}{4.2pt}
\renewcommand{\arraystretch}{0.92}
\begin{tabular*}{0.96\textwidth}{@{\extracolsep{\fill}} l | cc | cccccc}
\toprule
\multirow{3}{*}{\textbf{Generator}} 
& \multicolumn{2}{c|}{\textbf{Strategy Quality}}
& \multicolumn{6}{c}{\textbf{Strategy Utility}} \\
\cmidrule(lr){2-3}
\cmidrule(lr){4-9}
& \multirow{2}{*}{\textbf{Conc.}}
& \multirow{2}{*}{\textbf{Fmat.}}
& \multicolumn{2}{c}{\textbf{Exec. 1.7B}}
& \multicolumn{2}{c}{\textbf{Exec. 4B}}
& \multicolumn{2}{c}{\textbf{Exec. 8B}} \\
\cmidrule(lr){4-5}
\cmidrule(lr){6-7}
\cmidrule(lr){8-9}
& &
& \textbf{Corr.} & \textbf{Sta$_{\mathrm{samp}}$}
& \textbf{Corr.} & \textbf{Sta$_{\mathrm{samp}}$}
& \textbf{Corr.} & \textbf{Sta$_{\mathrm{samp}}$} \\
\midrule

\textbf{1.7B} 
& 90.98 & 34.88
& 14.50 & 48.30
& 18.48 & 54.97
& 15.86 & 54.49 \\

\textbf{4B}
& 90.03 & 25.12
& 13.27 & 42.44
& 19.41 & 53.66
& 14.96 & 53.98 \\

\textbf{8B}
& 91.57 & 22.09
& \textbf{14.57} & 45.03
& \textbf{20.86} & 56.83
& \textbf{17.74} & 54.30 \\

\textbf{14B}
& \textbf{92.98} & \textbf{38.84}
& 13.21 & \textbf{48.51}
& 20.83 & \textbf{57.51}
& 16.74 & \textbf{54.73} \\

\bottomrule
\end{tabular*}
\vspace{-0.5em}
\caption{Effect of generator and executor model scales. We compare four strategy generators and evaluate their generated strategies with three executors. \textbf{Bold} values indicate the best result under the same metric and executor setting.}
\label{tab:scale_strategy_generation}
\end{table*}

\subsection{Factors Affecting Strategy Generation and Application}

To answer \textbf{RQ2}, we further analyze factors that affect strategy-based ICL, including model scale, the number of few-shot examples, and strategy format constraints.

\subsubsection{Effect of Generator and Executor Scale}

We use Qwen3-1.7B, 4B, 8B, and 14B as strategy generators, and use Qwen3-1.7B, 4B, and 8B as strategy executors. We evaluate different generator-executor combinations to analyze the effect of model scale. Table~\ref{tab:scale_strategy_generation} reports the results.

Overall, model scale affects strategy quality and utility differently. With a fixed executor, Qwen3-14B produces the best text quality, whereas Qwen3-8B often yields stronger execution performance. With a fixed generator, Qwen3-4B achieves the best execution results under most settings, while scaling the executor to Qwen3-8B does not consistently improve correctness. These results suggest that larger generators improve strategy readability and format quality, but effective strategy application depends on a proper generator--executor match.





\subsubsection{Effect of Few-shot Number}

To analyze the effect of the number of examples, we compare the strategy quality and utility of strategy-based ICL under $1$-, $3$-, $5$-, $10$-, and $30$-shot settings. Table~\ref{tab:fewshot_strategy_generation} reports the results.

\begin{table}[!htbp]
\centering
\scriptsize
\setlength{\tabcolsep}{2.6pt}
\renewcommand{\arraystretch}{0.90}
\setlength{\aboverulesep}{0.20ex}
\setlength{\belowrulesep}{0.20ex}
\begin{tabular*}{\columnwidth}{@{\extracolsep{\fill}}lcccc}
\toprule
\multirow{2}{*}{\textbf{$k$}}
& \multicolumn{2}{c}{\textbf{Strategy Quality}}
& \multicolumn{2}{c}{\textbf{Strategy Utility}} \\
\cmidrule(lr){2-3}
\cmidrule(lr){4-5}
& \textbf{Conc.}
& \textbf{Fmat.}
& \textbf{Corr.}
& \textbf{Sta$_{\mathrm{samp}}$} \\
\midrule
1  & \textbf{92.67} & \textbf{31.33} & 17.34 & 46.51 \\
3  & 91.57 & 22.09 & 20.89 & 48.69 \\
5  & 91.54 & 21.83 & 21.26 & 48.06 \\
10 & 90.77 & 15.41 & 21.19 & 48.41 \\
30 & 89.83 & 14.58 & \textbf{25.49} & \textbf{51.29} \\
\bottomrule
\end{tabular*}
\vspace{-0.4em}
\caption{Effect of the number of few-shot examples on strategy generation and downstream application under $k=1,3,5,10,30$. \textbf{Bold} values indicate the best performance across different few-shot settings.}
\label{tab:fewshot_strategy_generation}
\end{table}


This is because as the number of examples increases, the generated strategy becomes more detailed. Although such a strategy may be less concise and could even exceed length limits and become unusable, it helps the model answer questions more effectively. This implies that the model may be better at leveraging detailed rather than concise strategies.

\subsubsection{Effect of Prompt Variant}

In addition to model scale and the number of examples, prompt design also affects
how models interpret the objective of strategy generation. We compare two prompt
variants: a free-form prompt and a structured prompt. The free-form prompt uses
flexible output constraints to elicit concise and executable strategies, whereas
the structured prompt imposes explicit field and format constraints to improve
output regularity and parsability. Their full prompt templates are provided in
Appendix~\ref{app:prompts}, and Table~\ref{tab:prompt_strategy_generation}
reports the comparison results.

\begin{table}[!htbp]
\centering
\scriptsize
\setlength{\tabcolsep}{4.2pt}
\renewcommand{\arraystretch}{0.98}
\setlength{\aboverulesep}{0.22ex}
\setlength{\belowrulesep}{0.22ex}
\resizebox{\columnwidth}{!}{
\begin{tabular}{lcccc}
\toprule
\multirow{2}{*}{\textbf{Prompt Type}} 
& \multicolumn{2}{c}{\textbf{Strategy Quality}} 
& \multicolumn{2}{c}{\textbf{Strategy Utility}} \\
\cmidrule(lr){2-3}
\cmidrule(lr){4-5}
& \textbf{Conc.} 
& \textbf{Fmat.}
& \textbf{Corr.}
& \textbf{Sta$_{\mathrm{samp}}$} \\
\midrule

Structured Prompt & \textbf{91.60} & 22.79 & \textbf{20.86} & \textbf{48.55} \\
Free-form Prompt  & 90.91 & \textbf{57.38} & 16.96 & 45.02 \\

\bottomrule
\end{tabular}
}
\caption{Comparison between structured and free-form prompts for strategy generation and downstream application. \textbf{Bold} values indicate the better result between the two prompt types under the same metric.}
\label{tab:prompt_strategy_generation}
\end{table}

The free-form prompt achieves higher Fmat., while the structured prompt performs
better on Corr. and Sta$_{\mathrm{samp}}$. This suggests that the free-form prompt
is more likely to produce outputs that satisfy the required format, whereas the
structured prompt yields strategies with stronger downstream utility and sampling
stability. Overall, prompt design affects strategy generation in different ways:
format compliance does not necessarily imply better downstream performance, and
the utility of a strategy also depends on whether it captures task-relevant rules
effectively.

\begin{table*}[t]
\centering
\scriptsize
\setlength{\tabcolsep}{3.2pt}
\renewcommand{\arraystretch}{1.10}
\resizebox{\textwidth}{!}{
\begin{tabular}{l | ccccc | ccccc | ccccc}
\toprule
\multirow{3}{*}{\textbf{Method}}
& \multicolumn{5}{c|}{\textbf{ID}}
& \multicolumn{5}{c|}{\textbf{OOD}}
& \multicolumn{5}{c}{\textbf{BBH}} \\
\cmidrule(lr){2-6}
\cmidrule(lr){7-11}
\cmidrule(lr){12-16}
& \multicolumn{2}{c}{\textbf{Quality}}
& \multicolumn{3}{c|}{\textbf{Utility}}
& \multicolumn{2}{c}{\textbf{Quality}}
& \multicolumn{3}{c|}{\textbf{Utility}}
& \multicolumn{2}{c}{\textbf{Quality}}
& \multicolumn{3}{c}{\textbf{Utility}} \\
\cmidrule(lr){2-3}
\cmidrule(lr){4-6}
\cmidrule(lr){7-8}
\cmidrule(lr){9-11}
\cmidrule(lr){12-13}
\cmidrule(lr){14-16}
& \textbf{Conc.}
& \textbf{Fmat.}
& \textbf{Corr.}
& \textbf{Sta$_{\mathrm{samp}}$}
& \textbf{Sta$_{\mathrm{rob}}$}
& \textbf{Conc.}
& \textbf{Fmat.}
& \textbf{Corr.}
& \textbf{Sta$_{\mathrm{samp}}$}
& \textbf{Sta$_{\mathrm{rob}}$}
& \textbf{Conc.}
& \textbf{Fmat.}
& \textbf{Corr.}
& \textbf{Sta$_{\mathrm{samp}}$}
& \textbf{Sta$_{\mathrm{rob}}$} \\
\midrule

Ours
& 92.13 & 6.68
& \textbf{22.71} & 59.61 & 68.08
& 91.46 & 21.94
& \textbf{25.42} & 59.67 & 68.18
& 91.77 & 5.15
& \textbf{15.34} & 51.90 & 62.32 \\

Ours+SFT
& \textbf{94.17} & \textbf{62.06}
& 21.33 & \textbf{70.08} & \textbf{77.79}
& \textbf{93.03} & \textbf{61.95}
& 23.29 & \textbf{62.89} & \textbf{72.58}
& \textbf{93.44} & \textbf{50.35}
& 14.20 & \textbf{66.51} & \textbf{76.41} \\

\bottomrule
\end{tabular}
}
\caption{Comparison between Ours and Ours+SFT on ID, OOD, and BBH test sets. Each split reports two strategy quality metrics and three strategy utility metrics. \textbf{Bold} values indicate the best performance between the two variants under the same evaluation setting.}
\label{tab:ours_sft_id_ood_bbh}
\end{table*}

\subsection{Effect of SFT on Strategy Generation Capability}

To answer \textbf{RQ3}, we compare the original Ours with the supervised fine-tuned variant Ours+SFT on ID, OOD, and BBH test sets. Table~\ref{tab:ours_sft_id_ood_bbh} reports the results.

Ours+SFT consistently improves Conc. and Fmat. over Ours on ID, OOD, and BBH. This indicates that SFT improves the conciseness and format compliance of generated strategies and generalizes across different test distributions. Ours+SFT also improves Sta$_{\mathrm{samp}}$ and Sta$_{\mathrm{rob}}$ on all three test sets, showing that SFT makes strategy generation more stable and robust.
On the other hand, although stability has improved, it remains far from 100\%, meaning the model retains a certain level of exploratory capability.

However, Ours+SFT does not improve Corr. and even leads to a slight decrease. This suggests that improving the form and stability of strategy generation does not necessarily translate into higher single-run answer correctness. Strategy utility is also affected by task difficulty and the ability of the executor to apply the generated strategy.

Therefore, although the SFT model does not become ``smarter'', it achieves greater stability and controllability while maintaining some exploratory ability. This suggests that it could serve as a strong initialization for reinforcement learning~\citep{ouyang2022rlhf}, which remains a direction for future work.

\section{Related Work}

\paragraph{Few-shot In-context Learning.}
Few-shot in-context learning (ICL) enables large language models to adapt to new tasks using only a small number of examples without parameter updates~\citep{brown2020language}. Prior studies show that ICL is highly sensitive to example selection, semantic relevance, and example order~\citep{liu2021makes,rubin2022learning,lu2022fantastically}. Increasing the number of examples does not fully eliminate the effects of context length and example organization~\citep{agarwal2024many}. To improve the task-solving ability of few-shot ICL, researchers have proposed several enhanced prompting paradigms, including chain-of-thought prompting~\citep{wei2022chain}, self-consistency~\citep{wang2023selfconsistency}, and least-to-most prompting~\citep{zhou2023least}. Recent studies further explore explicit intermediate structures, such as Self-Discover, PAL, and PoT~\citep{zhou2024selfdiscover,gao2023pal,chen2023program}. These methods improve performance in specific reasoning scenarios, but they usually serve individual queries or specific reasoning forms. In contrast, this paper focuses on whether models can induce transferable task-level strategies from few-shot examples.

\paragraph{Benchmarks for In-context Learning.}
Existing benchmarks mainly evaluate final answer accuracy or task scores, such as MMLU, BIG-Bench, BBH, and GSM8K~\citep{hendrycks2020measuring,srivastava2022beyond,suzgun2022challenging,cobbe2021training}. They provide limited evaluation of the reasoning process. Some recent studies begin to evaluate or supervise process-level reasoning, such as step verification for mathematical reasoning and reasoning trace evaluation~\citep{lightman2023lets,lee2025evaluating}. However, these evaluations usually focus on the solution process of individual examples, rather than whether models can induce general strategies that transfer to similar examples. This paper takes explicit strategy induction as the core evaluation target and further evaluates the executability of induced strategies through downstream task performance. Table~\ref{tab:benchmark_comparison} compares representative benchmarks and evaluations.

\paragraph{Concurrent work.}
A concurrent anonymous study uses \textsc{StrategyBench} and an SFT model trained from our reference strategies to develop CoDI, a downstream method for OOD generalization~\citep{anonymous2026codi}. 
This work is complementary to ours, which focuses on benchmark construction and evaluation; details are provided in Appendix~\ref{app:concurrent_work}.

\begin{table}[t]
\centering
\scriptsize
\setlength{\tabcolsep}{1.6pt}
\renewcommand{\arraystretch}{1.12}
\resizebox{\columnwidth}{!}{
\begin{tabular}{lcccc}
\toprule
\textbf{Benchmark / Eval.} 
& \makecell{\textbf{Ans.}\\\textbf{Eval.}} 
& \makecell{\textbf{Proc.}\\\textbf{Eval.}} 
& \makecell{\textbf{Strategy}\\\textbf{Ind.}} 
& \makecell{\textbf{Strategy}\\\textbf{Utility}} \\
\midrule
MMLU~\citep{hendrycks2020measuring} 
& \checkmark & $\times$ & $\times$ & $\times$ \\
BBH~\citep{suzgun2022challenging} 
& \checkmark & $\times$ & $\times$ & $\times$ \\
GSM8K~\citep{cobbe2021training} 
& \checkmark & $\times$ & $\times$ & $\times$ \\
Step Verif.~\citep{lightman2023lets} 
& \checkmark & \checkmark & $\times$ & $\times$ \\
Trace Eval.~\citep{lee2025evaluating} 
& \checkmark & \checkmark & $\times$ & $\times$ \\
\textbf{Ours} 
& \checkmark & \checkmark & \checkmark & \checkmark \\
\bottomrule
\end{tabular}
}
\caption{Comparison of representative benchmarks and process-level evaluations. 
Ans. Eval., Proc. Eval., Strategy Ind., and Strategy Utility denote answer-level 
evaluation, process-level evaluation, strategy induction, and strategy utility 
evaluation, respectively.}
\label{tab:benchmark_comparison}
\end{table}


\section{Conclusion}

This paper introduces \textsc{StrategyBench}, a benchmark for evaluating explicit strategy induction in large language models, together with metrics covering both strategy text quality and downstream application utility. Experiments show that explicit strategies can improve task-solving performance, but their utility depends on task type, example configuration, and the generator--executor match. Supervised fine-tuning further improves the stability and generalization of strategy generation.

Future work will extend \textsc{StrategyBench} to more task sources, including real-world, interactive, and tool-use scenarios. It is also important to distinguish strategy generation failures from execution failures more precisely and to explore training methods that enhance the transferability and executability of generated strategies.



\section{Limitations}

This paper has several limitations in methodology and experiments.

\textbf{Methodology.}
(1) \textsc{StrategyBench} is mainly built from BIG-Bench and BBH. Although the filtered tasks cover multiple reasoning categories, the data sources are still relatively limited. Future work can incorporate more independent benchmarks to evaluate the generalization of explicit strategy induction across broader task distributions.
(2) This paper evaluates generated strategies from two perspectives: strategy text quality and downstream application utility. However, it does not fully distinguish different sources of failure. For example, an incorrect answer may result from an inaccurate induced strategy, or from the executor's failure to understand and apply the strategy correctly. Future work should conduct a more fine-grained analysis of strategy generation errors and strategy execution errors.

\textbf{Experiments.}
(1) Due to computational resource constraints, this paper mainly evaluates models from the Qwen3 family. This setting enables controlled scale comparisons within the same model family, but it does not cover more model families or closed-source models.
(2) The fine-tuning experiments in this paper mainly focus on the effect of SFT on strategy generation capability. We have not further explored reinforcement learning, preference optimization, or multi-stage curriculum learning. These methods may further improve the executability and downstream utility of generated strategies.


\bibliography{custom}

@inproceedings{ouyang2022rlhf,
author = {Ouyang, Long and Wu, Jeff and Jiang, Xu and Almeida, Diogo and Wainwright, Carroll L. and Mishkin, Pamela and Zhang, Chong and Agarwal, Sandhini and Slama, Katarina and Ray, Alex and Schulman, John and Hilton, Jacob and Kelton, Fraser and Miller, Luke and Simens, Maddie and Askell, Amanda and Welinder, Peter and Christiano, Paul and Leike, Jan and Lowe, Ryan},
title = {Training language models to follow instructions with human feedback},
year = {2022},
isbn = {9781713871088},
publisher = {Curran Associates Inc.},
address = {Red Hook, NY, USA},
booktitle = {Proceedings of the 36th International Conference on Neural Information Processing Systems},
articleno = {2011},
numpages = {15},
location = {New Orleans, LA, USA},
series = {NIPS '22}
}

@inproceedings{brown2020language,
  title = {Language Models are Few-Shot Learners},
  author = {Brown, Tom B. and Mann, Benjamin and Ryder, Nick and Subbiah, Melanie and Kaplan, Jared and Dhariwal, Prafulla and Neelakantan, Arvind and Shyam, Pranav and Sastry, Girish and Askell, Amanda and Agarwal, Sandhini and Herbert-Voss, Ariel and Krueger, Gretchen and Henighan, Tom and Child, Rewon and Ramesh, Aditya and Ziegler, Daniel M. and Wu, Jeffrey and Winter, Clemens and Hesse, Christopher and Chen, Mark and Sigler, Eric and Litwin, Mateusz and Gray, Scott and Chess, Benjamin and Clark, Jack and Berner, Christopher and McCandlish, Sam and Radford, Alec and Sutskever, Ilya and Amodei, Dario},
  booktitle = {Advances in Neural Information Processing Systems},
  year = {2020}
}

@inproceedings{wei2022chain,
  title = {Chain-of-Thought Prompting Elicits Reasoning in Large Language Models},
  author = {Wei, Jason and Wang, Xuezhi and Schuurmans, Dale and Bosma, Maarten and Ichter, Brian and Xia, Fei and Chi, Ed H. and Le, Quoc V. and Zhou, Denny},
  booktitle = {Advances in Neural Information Processing Systems},
  year = {2022}
}

@inproceedings{wang2023selfconsistency,
  title = {Self-Consistency Improves Chain of Thought Reasoning in Language Models},
  author = {Wang, Xuezhi and Wei, Jason and Schuurmans, Dale and Le, Quoc V. and Chi, Ed H. and Narang, Sharan and Chowdhery, Aakanksha and Zhou, Denny},
  booktitle = {International Conference on Learning Representations},
  year = {2023}
}

@inproceedings{zhou2023least,
  title = {Least-to-Most Prompting Enables Complex Reasoning in Large Language Models},
  author = {Zhou, Denny and Sch{\"a}rli, Nathanael and Hou, Le and Wei, Jason and Scales, Nathan and Wang, Xuezhi and Schuurmans, Dale and Cui, Claire and Bousquet, Olivier and Le, Quoc V. and Chi, Ed H.},
  booktitle = {International Conference on Learning Representations},
  year = {2023}
}

@article{cobbe2021training,
  title = {Training Verifiers to Solve Math Word Problems},
  author = {Cobbe, Karl and Kosaraju, Vineet and Bavarian, Mohammad and Chen, Mark and Jun, Heewoo and Kaiser, Lukasz and Plappert, Matthias and Tworek, Jerry and Hilton, Jacob and Nakano, Reiichiro and Hesse, Christopher and Schulman, John},
  journal = {arXiv preprint arXiv:2110.14168},
  year = {2021}
}

@inproceedings{hendrycks2020measuring,
  title = {Measuring Massive Multitask Language Understanding},
  author = {Hendrycks, Dan and Burns, Collin and Basart, Steven and Zou, Andy and Mazeika, Mantas and Song, Dawn and Steinhardt, Jacob},
  booktitle = {International Conference on Learning Representations},
  year = {2021}
}

@article{srivastava2022beyond,
  title = {Beyond the Imitation Game: Quantifying and Extrapolating the Capabilities of Language Models},
  author = {Srivastava, Aarohi and Rastogi, Abhinav and Rao, Abhishek and Shoeb, Abu Awal Md and Abid, Abubakar and Fisch, Adam and Brown, Alex R. and Santoro, Adam and Gupta, Aditya and Garriga-Alonso, Adri{\`a} and Kluska, Agnieszka and others},
  journal = {Transactions on Machine Learning Research},
  year = {2023}
}

@inproceedings{suzgun2022challenging,
  title = {Challenging BIG-Bench Tasks and Whether Chain-of-Thought Can Solve Them},
  author = {Suzgun, Mirac and Scales, Nathan and Sch{\"a}rli, Nathanael and Gehrmann, Sebastian and Tay, Yi and Chung, Hyung Won and Chowdhery, Aakanksha and Le, Quoc V. and Chi, Ed H. and Zhou, Denny and Wei, Jason},
  booktitle = {Findings of the Association for Computational Linguistics: ACL},
  year = {2023}
}

@inproceedings{lightman2023lets,
  title = {Let's Verify Step by Step},
  author = {Lightman, Hunter and Kosaraju, Vineet and Burda, Yura and Edwards, Harrison and Baker, Bowen and Lee, Teddy and Leike, Jan and Schulman, John and Sutskever, Ilya and Cobbe, Karl},
  booktitle = {International Conference on Learning Representations},
  year = {2024}
}

@article{mirzadeh2024gsm,
  title = {GSM-Symbolic: Understanding the Limitations of Mathematical Reasoning in Large Language Models},
  author = {Mirzadeh, Iman and Alizadeh, Keivan and Shahrokhi, Hooman and Tuzel, Oncel and Bengio, Samy and Farajtabar, Mehrdad},
  journal = {arXiv preprint arXiv:2410.05229},
  year = {2024}
}

@article{zeng2023mrgsm8k,
  title = {MR-GSM8K: A Meta-Reasoning Benchmark for Large Language Model Evaluation},
  author = {Zeng, Zhiyuan and Chen, Peng and Liu, Sheng and Jiang, Haoming and Jia, Jia},
  journal = {arXiv preprint arXiv:2312.17080},
  year = {2023}
}

@inproceedings{honovich2023instruction,
  title = {Instruction Induction: From Few Examples to Natural Language Task Descriptions},
  author = {Honovich, Or and Shaham, Uri and Bowman, Samuel R. and Levy, Omer},
  booktitle = {Proceedings of the 61st Annual Meeting of the Association for Computational Linguistics},
  pages = {1935--1952},
  year = {2023}
}

@inproceedings{zheng2024take,
  title = {Take a Step Back: Evoking Reasoning via Abstraction in Large Language Models},
  author = {Zheng, Huaixiu Steven and Mishra, Swaroop and Chen, Xinyun and Cheng, Heng-Tze and Chi, Ed H. and Le, Quoc V. and Zhou, Denny},
  booktitle = {International Conference on Learning Representations},
  year = {2024}
}

@inproceedings{chen2024skill,
  title = {Induct-Learn: Short Phrase Prompting with Instruction Induction},
  author = {Chen, Ping-Chen and Wei, Szu-Lin and Huang, Hen-Hsen and Chen, Hsin-Hsi},
  booktitle = {Proceedings of the 2024 Conference on Empirical Methods in Natural Language Processing},
  pages = {5204--5231},
  year = {2024}
}

@article{liu2021makes,
  title = {What Makes Good In-Context Examples for GPT-3?},
  author = {Liu, Jiachang and Shen, Dinghan and Zhang, Yizhe and Dolan, Bill and Carin, Lawrence and Chen, Weizhu},
  journal = {arXiv preprint arXiv:2101.06804},
  year = {2021}
}

@inproceedings{lu2022fantastically,
  title = {Fantastically Ordered Prompts and Where to Find Them: Overcoming Few-Shot Prompt Order Sensitivity},
  author = {Lu, Yao and Bartolo, Max and Moore, Alastair and Riedel, Sebastian and Stenetorp, Pontus},
  booktitle = {Proceedings of the 60th Annual Meeting of the Association for Computational Linguistics},
  year = {2022}
}

@inproceedings{rubin2022learning,
  title = {Learning To Retrieve Prompts for In-Context Learning},
  author = {Rubin, Ohad and Herzig, Jonathan and Berant, Jonathan},
  booktitle = {Proceedings of the 2022 Conference of the North American Chapter of the Association for Computational Linguistics: Human Language Technologies},
  year = {2022}
}

@article{agarwal2024many,
  title = {Many-Shot In-Context Learning},
  author = {Agarwal, Rishabh and Singh, Avi and Zhang, Lei M. and Bohnet, Bernd and Rosias, Lorena and Chan, Stephanie and Zhang, Biao and Anand, Ankesh and Abbas, Zaheer and Nova, Azade and Co-Reyes, John D. and Chu, Eric and Behbahani, Feryal and Faust, Aleksandra and Larochelle, Hugo},
  journal = {arXiv preprint arXiv:2404.11018},
  year = {2024}
}

@inproceedings{min2022rethinking,
  title = {Rethinking the Role of Demonstrations: What Makes In-Context Learning Work?},
  author = {Min, Sewon and Lyu, Xinxi and Holtzman, Ari and Artetxe, Mikel and Lewis, Mike and Hajishirzi, Hannaneh and Zettlemoyer, Luke},
  booktitle = {Proceedings of the 2022 Conference on Empirical Methods in Natural Language Processing},
  pages = {11048--11064},
  year = {2022},
  publisher = {Association for Computational Linguistics}
}

@article{liu2024lost,
  title = {Lost in the Middle: How Language Models Use Long Contexts},
  author = {Liu, Nelson F. and Lin, Kevin and Hewitt, John and Paranjape, Ashwin and Bevilacqua, Michele and Petroni, Fabio and Liang, Percy},
  journal = {Transactions of the Association for Computational Linguistics},
  volume = {12},
  pages = {157--173},
  year = {2024},
  publisher = {MIT Press}
}

@inproceedings{zhou2024selfdiscover,
  title={Self-Discover: Large Language Models Self-Compose Reasoning Structures},
  author={Zhou, Pei and Pujara, Jay and Ren, Xiang and Chen, Xinyun and Cheng, Heng-Tze and Le, Quoc V. and Chi, Ed H. and Zhou, Denny and Mishra, Swaroop and Zheng, Huaixiu Steven},
  booktitle={Advances in Neural Information Processing Systems},
  year={2024}
}

@inproceedings{gao2023pal,
  title={PAL: Program-aided Language Models},
  author={Gao, Luyu and Madaan, Aman and Zhou, Shuyan and Alon, Uri and Liu, Pengfei and Yang, Yiming and Callan, Jamie and Neubig, Graham},
  booktitle={Proceedings of the 40th International Conference on Machine Learning},
  year={2023}
}

@article{chen2023program,
  title={Program of Thoughts Prompting: Disentangling Computation from Reasoning for Numerical Reasoning Tasks},
  author={Chen, Wenhu and Ma, Xueguang and Wang, Xinyi and Cohen, William W.},
  journal={Transactions on Machine Learning Research},
  year={2023}
}

@inproceedings{lee2025evaluating,
  title={Evaluating Step-by-step Reasoning Traces: A Survey},
  author={Lee, Jinu and Hockenmaier, Julia},
  booktitle={Findings of the Association for Computational Linguistics: EMNLP 2025},
  pages={1789--1814},
  year={2025}
}

@inproceedings{kojima2022large,
  title     = {Large Language Models are Zero-Shot Reasoners},
  author    = {Kojima, Takeshi and Gu, Shixiang Shane and Reid, Machel and Matsuo, Yutaka and Iwasawa, Yusuke},
  booktitle = {Advances in Neural Information Processing Systems},
  volume    = {35},
  pages     = {22199--22213},
  year      = {2022}
}

@misc{anonymous2026codi,
  author = {Anonymous},
  title = {Cognitively Decoupled Meta-Induction for Out-of-Distribution Generalization},
  year = {2026},
  note = {Under review}
}





\appendix

\section{Task Selection and Standardization Details}
\label{app:task_selection}

This appendix provides additional details on the task selection and format standardization process of \textsc{StrategyBench}. Overall, we first follow the BBH-style procedure to preprocess and filter the original BIG-Bench tasks, removing tasks that are not suitable for unified few-shot strategy induction evaluation. We then adopt a ``model-assisted pre-filtering + human verification'' procedure to further select tasks suitable for strategy induction. Finally, we standardize the input and target formats of the retained examples.

\subsection{Task Filtering}
\label{app:task_filtering}

We first perform an initial filtering over the original BIG-Bench tasks and remove three types of tasks that are clearly unsuitable for static few-shot strategy induction evaluation. The first type includes tasks that rely on dynamic generation or external program execution for evaluation, as these tasks do not provide fixed input-output examples. The second type includes highly subjective or sensitive tasks, such as bias, toxicity, and ethical judgment tasks, which usually lack a single stable and objective gold answer. The third type includes examples whose input length exceeds 2500 characters. We remove these examples to reduce the interference of long contexts with few-shot example construction, strategy generation, and strategy application, and to control inference cost.

\subsection{Model-assisted Task Selection}
\label{app:model_assisted_selection}

After the initial filtering, we further determine whether each candidate subtask is suitable for explicit strategy induction evaluation. Specifically, we use Qwen3-32B to automatically score each candidate subtask, and then conduct human verification to finalize the selection. The scoring dimensions include objective scorability, format stability, rule inductibility, external knowledge risk, and strategy transfer potential. Each dimension is scored from 0 to 2. For all dimensions except external knowledge risk, a higher score indicates that the task is more suitable for strategy induction. External knowledge risk is a negative indicator, where a higher score means that the task relies more heavily on external knowledge.

\paragraph{Objective scorability.}
This dimension measures whether a task has clear and stable objective evaluation criteria. If a task can be evaluated by exact match or clear multiple-choice answers, it receives a score of 2. If its answers are partially objective but contain some ambiguity, it receives a score of 1. If the task is open-ended or subjective and has unclear answer boundaries, it receives a score of 0.

\paragraph{Format stability.}
This dimension measures whether the input-output format is stable within the same task or subtask. If the task format is fixed and the input-output structure is clear, it receives a score of 2. If the overall format is relatively stable but contains some noise or variants, it receives a score of 1. If the task format varies substantially and makes it difficult to extract stable patterns, it receives a score of 0.

\paragraph{Rule inductibility.}
This dimension measures whether a task contains rules, steps, or algorithms that can be abstracted from a small number of examples. If a task has clear inductive rules and can be written as stable operation steps, it receives a score of 2. If it contains certain patterns but the rules are incomplete or weakly generalizable, it receives a score of 1. If the task mainly relies on commonsense memory, linguistic intuition, or cultural background and is difficult to abstract into a general strategy, it receives a score of 0.

\paragraph{External knowledge risk.}
This dimension measures whether solving a task depends on the model's prior world knowledge, cultural knowledge, or domain-specific knowledge, rather than information that can be induced from few-shot examples. This dimension is a negative indicator. If a task heavily depends on external knowledge, it receives a score of 2. If it requires some external knowledge but external knowledge is not the dominant factor, it receives a score of 1. If the task can mostly be solved through examples and induced rules, it receives a score of 0.

\paragraph{Strategy utility potential.}
This dimension measures whether an explicit strategy can potentially bring practical gains compared with the no-strategy setting. For example, BIG-Bench tasks such as moral permissibility and simple ethical questions usually depend on value judgments, contextual understanding, or safety preferences, and are difficult to formulate as stable objective rules. Explicit strategies may oversimplify complex contexts into partial judgment rules and reduce the model's adaptability to specific examples. Therefore, these tasks have low strategy utility potential. If errors in a task can be substantially reduced by a clear strategy, the task receives a score of 2. If a strategy may provide some help but the improvement is limited or unstable across examples, it receives a score of 1. If a strategy is unlikely to improve performance, or the task mainly relies on memorization, commonsense knowledge, or surface matching, it receives a score of 0.

\subsection{Selection Rule}
\label{app:selection_rule}

Based on the five dimensions above, we define the suitability score of a candidate subtask as:
\begin{equation}
\mathrm{Score}
=
s_{\mathrm{obj}}
+
s_{\mathrm{format}}
+
s_{\mathrm{rule}}
+
s_{\mathrm{transfer}}
-
s_{\mathrm{know}},
\end{equation}
where $s_{\mathrm{obj}}$, $s_{\mathrm{format}}$, $s_{\mathrm{rule}}$, and $s_{\mathrm{transfer}}$ denote the scores of objective scorability, format stability, rule inductibility, and strategy transfer potential, respectively. $s_{\mathrm{know}}$ denotes the score of external knowledge risk.

Using only the total score may retain some tasks that are not suitable for strategy induction evaluation. For example, some tasks are format-stable and easy to score, but heavily rely on external knowledge or cultural background. Their performance changes may mainly come from the model's memorized knowledge rather than strategies induced from few-shot examples. To avoid this issue, we use a filtering rule based on both the total score and necessary constraints. A candidate subtask must satisfy:
\begin{equation}
\mathrm{Score} \ge 7,
\end{equation}
and:
\begin{equation}
s_{\mathrm{know}} \le 1,\quad
s_{\mathrm{obj}} = 2,\quad
s_{\mathrm{transfer}} = 2.
\end{equation}

Here, $s_{\mathrm{know}} \le 1$ controls the risk of external knowledge dependence, $s_{\mathrm{obj}} = 2$ ensures stable objective evaluation criteria, and $s_{\mathrm{transfer}} = 2$ ensures clear strategy transfer potential. After model-assisted scoring, we further conduct human verification and remove tasks that clearly deviate from our research goal, such as multimodal tasks, highly sensitive tasks, and tasks that are clearly inconsistent with rule induction and strategy transfer.

\subsection{Input and Target Standardization}
\label{app:standardization}

After task filtering, we standardize the retained examples so that different BIG-Bench tasks can be uniformly used for few-shot strategy induction and automatic evaluation. The standardization process mainly includes input field standardization and target answer standardization.

\paragraph{Input standardization.}
The input information in the original BIG-Bench examples may be scattered across multiple fields, including task prefix, example input prefix, input prefix, and example output prefix. We integrate these fields into a unified input field. Specifically, task prefix usually represents task instructions, input represents the concrete question text, choice prefix provides the option prefix for multiple-choice tasks, and example output prefix specifies the expected output format.

For multiple-choice examples, if the original options do not have explicit labels, we assign unified option IDs such as A/B/C/D. This ensures that model outputs can be aligned with automatic evaluation. This processing enables different tasks to be organized into a unified few-shot prompt format and supports subsequent exact-match evaluation.

\paragraph{Target standardization.}
We convert the original answers into a unified target field. For binary classification tasks, such as yes/no questions, we directly retain the answer word. For general multiple-choice tasks, we convert the answer into the corresponding option ID, such as A, B, C, or D. For tasks with too many options or unstable option labels, we directly retain the original answer string. For exact-match tasks, we retain the gold answer in string form.

\paragraph{Standardized instance format.}
After standardization, each example is converted into a unified format. A simplified example is shown below:
\begin{table}[t]
\centering
\small
\setlength{\tabcolsep}{4pt}
\renewcommand{\arraystretch}{1.12}
\begin{tabular}{p{0.22\columnwidth}p{0.68\columnwidth}}
\toprule
\textbf{Field} & \textbf{Example Value} \\
\midrule
\texttt{task} 
& \texttt{disambiguation\_qa} \\

\texttt{subtask} 
& \texttt{disambiguation\_qa} \\

\texttt{input} 
& In the following sentences, explain the antecedent of the pronoun. Options: (A) \ldots{} (B) \ldots{} (C) Ambiguous. \\

\texttt{target} 
& \texttt{["A"]} \\
\bottomrule
\end{tabular}
\caption{Example of a standardized instance.}
\label{tab:standardized_instance_example}
\end{table}

Here, task denotes the original BIG-Bench task name, subtask denotes the corresponding subtask, input denotes the normalized model input, and target denotes the gold answer list used for automatic evaluation.

\section{Dataset Splitting and Sampling Details}
\label{app:dataset_splitting}

This section provides additional details on the dataset splitting and sampling rules of \textsc{StrategyBench}. Since the original BIG-Bench tasks vary substantially in sample size, subtask scale, input-output format, and evaluation difficulty, directly using all examples may cause large-scale tasks to dominate training or evaluation and reduce comparability across tasks. Therefore, we follow two main principles when constructing the benchmark. First, we avoid obvious information leakage between the training and test sets. Second, we aim to keep the number of examples relatively balanced across tasks and subtasks.

\paragraph{Training and test split.}
We first split the preprocessed BIG-Bench data into training and test sets. The training set is mainly used to construct few-shot examples, generate reference strategy data, and support model development and debugging. The test set is used to evaluate the strategy induction and strategy application abilities of models under different distribution conditions. To systematically analyze model generalization, we further divide the test set into in-distribution and out-of-distribution subsets.

\paragraph{In-distribution test set.}
The in-distribution test set is denoted as ID. Test examples in this set come from the same subtask set as the training examples, but the concrete examples do not overlap. This setting evaluates whether models can induce stable and effective strategies from few-shot examples and apply them to new test examples when the task rules and input-output formats are relatively consistent.

\paragraph{Out-of-distribution test set.}
The out-of-distribution test set is used to evaluate strategy induction and transfer ability beyond the training distribution. We assign examples to the OOD test set when they differ clearly from the training set in task source, input-output format, or required ability. We also avoid direct overlap with task rules and example templates used during training. This setting examines whether models can induce transferable explicit strategies from few-shot examples when facing unseen task structures or distribution shifts.

\paragraph{BBH test set.}
In addition to the internal BIG-Bench split, we retain BIG-Bench Hard (BBH) as an additional challenging test set. BBH contains a set of tasks that are more challenging for large language models and can be used to examine strategy induction ability in complex reasoning, symbolic manipulation, and multi-step problem-solving scenarios. We select tasks from BBH that are suitable for automatic evaluation and strategy induction, and ensure that they do not directly overlap with tasks in the training set. The final retained BBH set serves as an additional challenging out-of-distribution evaluation set.

\paragraph{Balanced sampling strategy.}
Since different tasks and subtasks vary greatly in sample size, we use a $k$-cut based balanced sampling method when constructing the training and evaluation sets. Given a set of data collections $\mathcal{D}_1,\mathcal{D}_2,\ldots,\mathcal{D}_m$, where $|\mathcal{D}_i|$ denotes the number of examples in the $i$-th collection, the sampling goal is to retain at most $k$ examples from each collection. If a collection contains fewer than $k$ examples, all examples in that collection are retained. To ensure that the overall sample size is not smaller than a predefined target size $N$, we solve for the smallest integer $k$ that satisfies:
\begin{equation}
    f(k) = \sum_{i=1}^{m} \min(k, |\mathcal{D}_i|) \geq N .
\end{equation}

Here, $f(k)$ denotes the total number of examples obtained when at most $k$ examples are sampled from each collection. Since $f(k)$ is monotonically increasing with respect to $k$, the smallest feasible $k$ can be efficiently found by binary search. After obtaining $k$, we apply the following sampling rule to each collection: if $|\mathcal{D}_i| \leq k$, all examples are retained; if $|\mathcal{D}_i| > k$, we randomly sample $k$ examples from the collection. This method controls the overall sample size while preventing a small number of large-scale tasks from contributing too many examples, making the influence of different tasks more balanced during training and evaluation.

\paragraph{Boundary cases.}
We handle boundary cases as follows. If the total number of examples across all candidate collections is still smaller than the target size, i.e.,
\begin{equation}
    \sum_{i=1}^{m} |\mathcal{D}_i| \leq N ,
\end{equation}
we do not perform truncation and directly retain all examples. If the number of candidate collections is already no smaller than the target size, i.e., $m \geq N$, we set $k=1$ to ensure that each collection contributes at least one example and preserve task coverage as much as possible. In other cases, we solve for the smallest $k$ that satisfies $f(k) \geq N$ and then perform sampling as described above.

\section{Task Categorization}
\label{app:task_categorization}

To support category-aware strategy generation and task-type-specific result analysis, we further divide the retained tasks into six coarse-grained semantic categories. This categorization does not directly follow the original BIG-Bench task labels. Instead, it considers the core abilities required by each task, the input-output format, and the underlying solving process. Compared with the original task names, our categorization focuses more on the types of strategies that models need to induce and execute when solving the tasks. Therefore, it is more suitable for analyzing how explicit strategies affect different task types.

\paragraph{Category definitions.}
We define the six coarse-grained task categories as follows:

\begin{itemize}
    \item \textbf{Numerical.}
    This category mainly involves numerical computation, arithmetic operations, counting, modular arithmetic, quantity comparison, or other rules related to numerical manipulation. Models usually need to perform deterministic mathematical or symbolic computations based on the numerical information in the question.

    \item \textbf{Logic.}
    This category mainly evaluates logical reasoning, rule constraints, symbolic relations, and consistency judgments. Models need to identify the conditions or constraints in the question and follow specific reasoning steps to obtain the answer.

    \item \textbf{Language.}
    This category focuses on natural language understanding, including lexical relations, syntactic patterns, textual entailment, paraphrase recognition, and judgments based on linguistic cues. The key ability is to extract semantic relations from text and make predictions accordingly.

    \item \textbf{Spatial.}
    This category involves spatial positions, directional relations, object movement, state changes, or transformations in structured spaces. Models usually need to maintain intermediate spatial states and reason based on changes in positions or directions.

    \item \textbf{Procedural.}
    This category requires models to follow given steps, operation sequences, or procedural rules. The main challenge is to correctly track intermediate states during multi-step operations and avoid state update errors in long-step reasoning.

    \item \textbf{Induction.}
    This category requires models to induce hidden rules, mapping relations, or abstract patterns from examples and transfer the induced rules to new query examples. Since these tasks explicitly rely on the process of summarizing rules from examples and then applying them, they are highly related to the strategy induction ability studied in this paper.
\end{itemize}

\paragraph{Categorization procedure.}
During categorization, we first construct a textual representation for each task. This representation consists of standardized input examples, output formats, and task-related metadata. We then use a semantic embedding model to encode the task representations and perform unsupervised clustering based on the embedding vectors to obtain initial task groups. Since automatic clustering may be affected by surface textual forms and place tasks with different solving mechanisms into the same group, we further conduct human semantic verification. Based on the main ability requirements and solving structures of the tasks, we merge the initial clusters into the six coarse-grained categories defined above. This process ensures that tasks within the same category have similar strategy requirements, while different categories reflect clear differences in reasoning abilities.

\section{Reference Strategy Construction Details}
\label{app:strategy_construction}

This section provides additional details on the construction of the reference strategy sets. We do not manually annotate solution processes for each test example. Instead, we automatically generate candidate strategies that summarize task-solving rules from few-shot examples, and then filter effective reference strategies based on downstream execution performance. The overall process consists of three steps: category-aware prompt design, candidate strategy generation, and utility-driven strategy filtering.

\paragraph{Category-aware prompt design.}
Different task types rely on different solving rules. For example, numerical tasks usually require step-by-step computation and boundary condition checking. Logic tasks rely more on constraint identification and stepwise reasoning. Procedural tasks require maintaining intermediate states, while induction tasks require extracting reusable hidden rules from examples. Therefore, we do not use a single generic prompt to generate all candidate strategies. Instead, we design category-aware prompts for strategy generation according to task categories.

Specifically, we design strategy generation prompts for the six coarse-grained task categories. Each category has two generation templates. One template emphasizes step-by-step reasoning and executable operation procedures, while the other emphasizes concise summaries of core rules. The former is suitable for tasks that require explicit state tracking, constraint checking, or multi-step reasoning. The latter is suitable for tasks with short rules, stable patterns, or relatively direct solving processes. In total, we obtain 12 category-aware strategy generation prompts to cover the strategy expression needs of different task types.

\paragraph{Candidate strategy generation.}
Given the few-shot example set $\mathcal{D}_{\mathrm{demo}}$ for a task or subtask, we use a strategy generator $f_{\theta_g}$ to generate candidate strategies:
\begin{equation}
    s = f_{\theta_g}(\mathcal{D}_{\mathrm{demo}}, p_c),
\end{equation}
where $p_c$ denotes the category-aware prompt corresponding to the current task category $c$, and $s$ denotes the generated candidate strategy. A candidate strategy may take the form of natural language rules, operation steps, input-output mapping relations, constraint checking methods, or error avoidance instructions. Its goal is not to restate the few-shot examples, but to extract general solving rules that can transfer to new query examples.

For each few-shot setting used to construct strategies, we select the corresponding prompt according to the task category and feed the few-shot examples into the strategy generator to obtain candidate strategies. The generated candidate strategies are not directly used as final reference strategies. Instead, they are passed to the downstream utility filtering stage.

\paragraph{Utility-based strategy filtering.}
Automatically generated candidate strategies may contain redundancy, errors, or information that is not helpful for downstream answering. Therefore, we further filter candidate strategies based on their execution performance on unseen query examples. Specifically, we use Direct ICL as the baseline and add the candidate strategy into the context to form strategy-based ICL. We then compare the accuracy of the two methods on the corresponding unseen query set $\mathcal{D}_{\mathrm{query}}$.

Let $\mathrm{Acc}_{\mathrm{direct}}$ denote the accuracy of Direct ICL, and let $\mathrm{Acc}_{\mathrm{strategy}}(s)$ denote the accuracy after using the candidate strategy $s$. If the candidate strategy satisfies:
\begin{equation}
    \mathrm{Acc}_{\mathrm{strategy}}(s) >
    \mathrm{Acc}_{\mathrm{direct}},
\end{equation}
we consider the strategy to bring downstream performance gains and add it to the effective strategy set $\mathcal{S}_{\mathrm{eff}}$.

Based on the effective strategy set, we further select strategies with more stable gains and obtain the significant reference strategy set $\mathcal{S}_{\mathrm{sig}}$:
\begin{equation}
    \mathcal{S}_{\mathrm{sig}} \subseteq \mathcal{S}_{\mathrm{eff}} .
\end{equation}
Here, $\mathcal{S}_{\mathrm{eff}}$ retains all candidate strategies that improve performance over Direct ICL, while $\mathcal{S}_{\mathrm{sig}}$ further emphasizes the stability and reliability of strategy gains. We use $\mathcal{S}_{\mathrm{sig}}$ as the main high-quality reference strategy set and use $\mathcal{S}_{\mathrm{eff}}$ for supplementary statistics and analysis.

\paragraph{Resulting strategy sets.}
Finally, we construct two levels of reference strategy data: the effective strategy set $\mathcal{S}_{\mathrm{eff}}$ and the significant reference strategy set $\mathcal{S}_{\mathrm{sig}}$. The former characterizes the range of automatically generated strategies that bring practical gains, while the latter provides a higher-quality and more stable source of reference strategies. The overall statistics of the two strategy sets are shown in Table~\ref{tab:strategy_set_statistics} in the main paper. 

\section{Implementation Details}
\label{app:implementation_details}

This section provides additional details on strategy generation, strategy application, and supervised fine-tuning settings. The main paper only reports the major experimental settings, while the detailed implementation parameters are summarized in this section.

\paragraph{Strategy generation settings.}
We use a local OpenAI-compatible inference interface for strategy generation. Unless otherwise specified, strategy generation uses prompt v1 as the prompt template, the default number of few-shot examples is $k=3$, and each bucket generates one candidate strategy by default. The generation temperature is set to $0.2$, and the maximum generation length is set to $700$ tokens. To improve request stability, we allow at most $6$ retries and set a $0.2$-second interval between adjacent requests. For stability and example perturbation experiments, we construct $5$ strategy variants for each bucket, including $3$ generations with the original example order and $2$ generations with randomly permuted example orders. The random seed is set to $42$.

\paragraph{Strategy application settings.}
During strategy application, the executor model receives the induced strategy and a new query question, and is instructed to output only the final answer. Unless otherwise specified, the decoding temperature in the strategy application stage is set to $0.7$, the maximum output length is set to $256$ tokens, and at most $3$ retries are allowed. The inference process supports concurrent requests, with the default concurrency set to $8$. For test example loading, the script prioritizes the test20 json file under the corresponding task directory. If full testing is enabled, all test examples in the file are evaluated. For answer judgment, we first extract the final answer from the model output, apply lightweight normalization, and then compare it with the gold answer using exact match.

\paragraph{Supervised fine-tuning settings.}
For SFT experiments, we construct supervised fine-tuning data from few-shot examples and their reference strategies, and use LoRA for parameter-efficient fine-tuning. The training model is Qwen3-8B. The maximum sequence length is set to $4096$, and the maximum number of training steps is set to $2200$. The per-device batch size is $1$, and the gradient accumulation step is $8$. The learning rate is set to $1\times10^{-4}$, and the warmup ratio is set to $0.03$. The model is evaluated and saved every $100$ steps, and training logs are recorded every $10$ steps. The LoRA rank is set to $16$, LoRA alpha is set to $32$, the target modules are attention layers, and training uses bf16 precision.

\begin{table}[t]
\centering
\small
\begin{tabular}{lc}
\toprule
\textbf{Hyperparameter} & \textbf{Value} \\
\midrule
Base model & Qwen3-8B \\
Max length & 4096 \\
Max steps & 2200 \\
Batch size & 1 \\
Gradient accumulation & 8 \\
Learning rate & $1\times10^{-4}$ \\
Warmup ratio & 0.03 \\
Eval / Save steps & 100 / 100 \\
Logging steps & 10 \\
LoRA rank & 16 \\
LoRA alpha & 32 \\
Target modules & Attention layers \\
Precision & bf16 \\
\bottomrule
\end{tabular}
\caption{Main hyperparameter settings for supervised fine-tuning.}
\label{tab:sft_hyperparameters}
\end{table}

\begin{figure*}[p]
\centering
\includegraphics[width=0.95\textwidth,height=0.82\textheight,keepaspectratio]{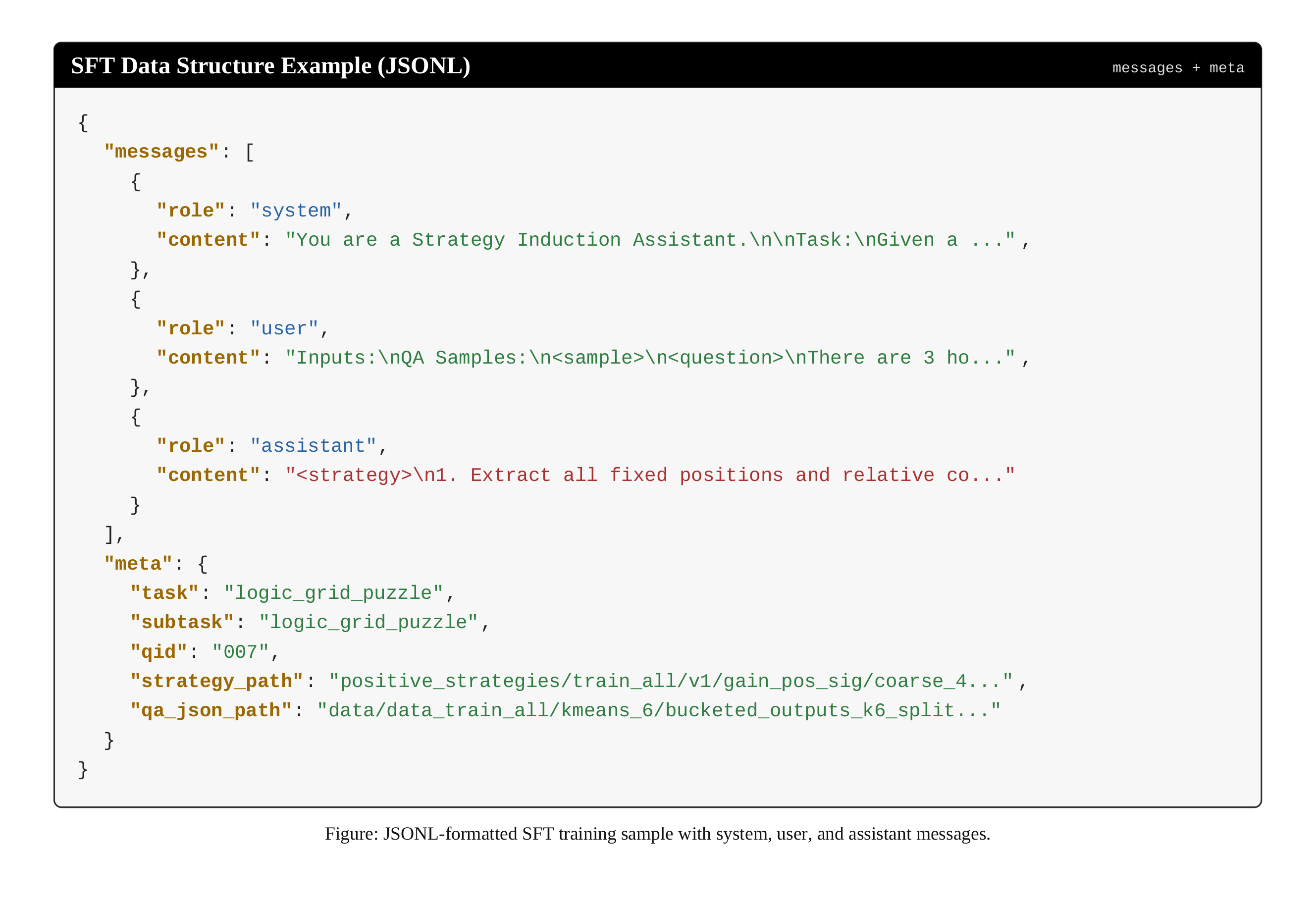}
\caption{Example of the JSONL-style data structure used for strategy supervised fine-tuning. Each instance contains the strategy-generation instruction, the few-shot QA block, and the reference strategy as the supervised target.}
\label{fig:sft_jsonl_example}
\end{figure*}

\paragraph{SFT data format.}
Each SFT instance follows the previous JSONL message style. The input contains the strategy-generation instruction and the few-shot QA block, while the output is the corresponding reference strategy used as the supervised target. Figure~\ref{fig:sft_jsonl_example} shows an example of the SFT data structure.

\paragraph{Use of AI Writing Assistance.}
We used GPT-based writing assistance solely to polish the language of author-written text, including grammar correction, wording refinement, and clarity improvement. The tool was not used to generate research ideas, experimental results, analyses, citations, or conclusions. All content was reviewed, revised, and verified by the authors, who take full responsibility for the final manuscript.

\section{Prompt Templates}
\label{app:prompts}

This section provides the prompt templates used for strategy generation. 
We consider two types of prompts: a structured prompt and a free-form prompt. 
The structured prompt specifies explicit output fields and format constraints, aiming to make the generated strategies more regular and easier to parse. 
The free-form prompt imposes fewer structural constraints and instead encourages the model to summarize concise and executable task-level strategies from the demonstrations. 
In this paper, Prompt V1 refers to the structured prompt, and Prompt V2 refers to the free-form prompt. 
Unless otherwise specified, Prompt V1 is used as the default strategy generation prompt in the main experiments. 

\begin{figure*}[p]
\centering
\includegraphics[
  width=0.95\textwidth,
  height=0.85\textheight,
  keepaspectratio
]{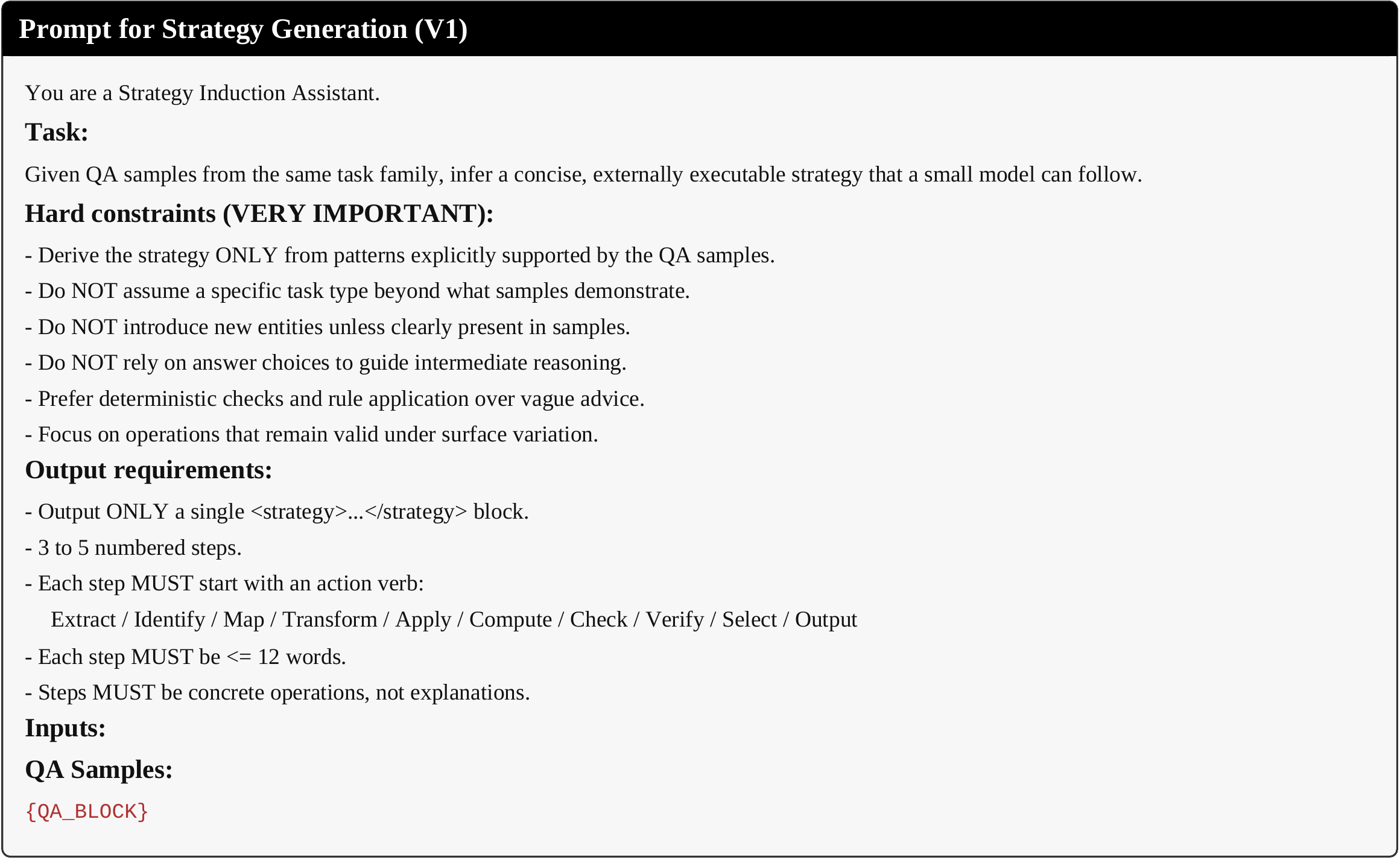}
\caption{Prompt V1 used for strategy generation.}
\label{fig:prompt_v1}
\end{figure*}

\begin{figure*}[p]
\centering
\includegraphics[
  width=0.95\textwidth,
  height=0.85\textheight,
  keepaspectratio
]{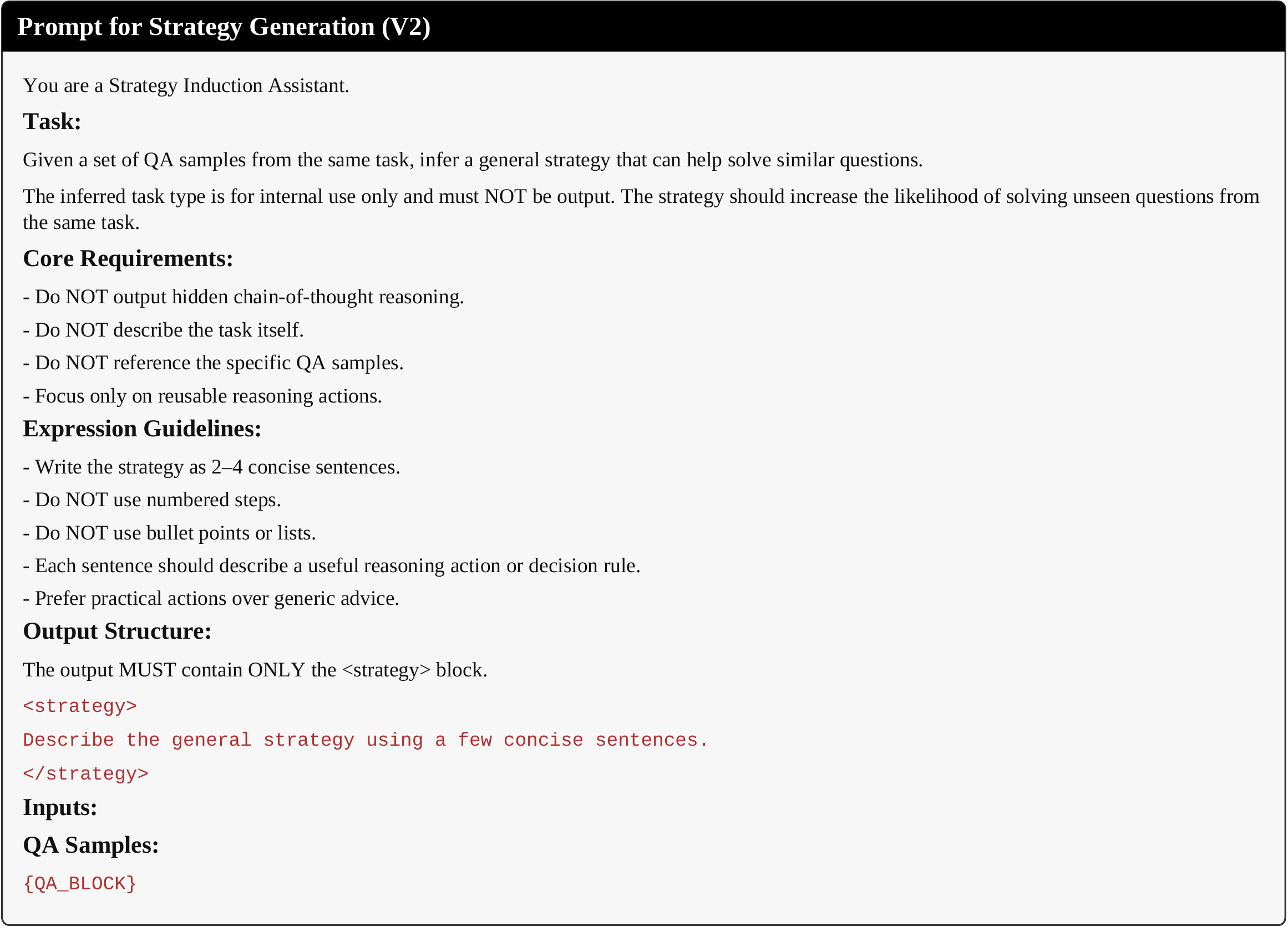}
\caption{Prompt V2 used for strategy generation.}
\label{fig:prompt_v2}
\end{figure*}

\section{Concurrent Work}
\label{app:concurrent_work}

A concurrent anonymous study proposes CoDI for cognitively decoupled meta-induction and OOD generalization~\citep{anonymous2026codi}. 
It uses \textsc{StrategyBench} as a training and evaluation testbed and further leverages an SFT model trained from our reference strategies. 
The concurrent study develops a downstream method based on these resources, whereas this paper focuses on constructing the benchmark, reference strategy resources, and evaluation framework for explicit strategy induction.

\begin{figure*}[t]
\centering
\includegraphics[width=0.92\textwidth]{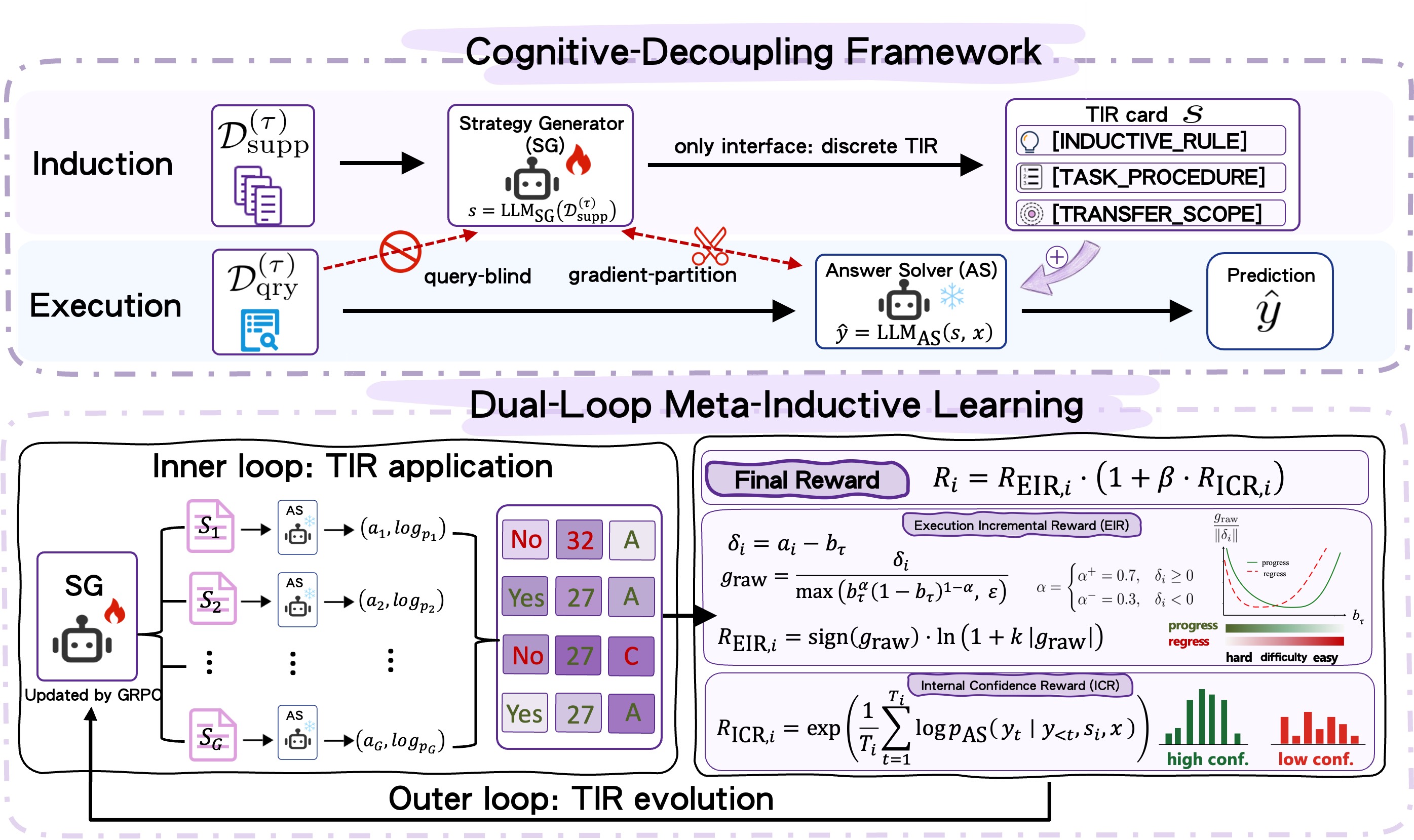}
\caption{Overview of the concurrent CoDI framework. CoDI uses \textsc{StrategyBench} as a training and evaluation testbed and leverages an SFT model trained from reference strategies. The framework separates task-level strategy induction from answer execution, serving as a downstream method built on the benchmark resources introduced in this paper.}
\label{fig:concurrent_codi}
\end{figure*}

Figure~\ref{fig:concurrent_codi} illustrates the overall framework of CoDI. 
The method separates strategy induction from answer execution: a strategy generator first induces task-level representations from support examples, and a frozen answer solver then applies the induced representation to answer new queries. 
This design is complementary to our benchmark-oriented contribution.

\end{document}